\pdfoutput=1

\documentclass[11pt]{article}
\usepackage{naacl2021}
\usepackage{times}
\usepackage{latexsym}

\usepackage{microtype}

\usepackage{caption}
\usepackage{subcaption}
\usepackage{amsmath}
\usepackage{amssymb}
\usepackage{amsfonts}
\usepackage{xfrac}
\usepackage{bbm}
\usepackage{amsthm}
\usepackage{graphicx}
\usepackage[T5,T1]{fontenc}
\usepackage{combelow}

\DeclareTextSymbolDefault{\ohorn}{T5}
\DeclareTextSymbolDefault{\uhorn}{T5}

\usepackage{booktabs}
\usepackage{cleveref}
\crefname{section}{\S}{\S\S}
\Crefname{section}{\S}{\S\S}
\crefname{table}{Table}{}
\crefname{figure}{Figure}{}
\crefname{algorithm}{Algorithm}{}
\crefname{equation}{eq.}{eqs.}
\crefname{appendix}{App.}{}
\crefname{prop}{Prop.}{}
\crefformat{section}{\S#2#1#3}  
\newcommand{\citeposs}[1]{\citeauthor{#1}'s (\citeyear{#1})}

\newcommand{\defn}[1]{\textbf{#1}}
\newcommand{\vw}{\mathbf{w}}

\newcommand{\dB}{d_{\mathbf{B}}}

\newcommand{\vt}{\mathbf{t}}
\newcommand{\vh}{\mathbf{h}}
\newcommand{\xx}{\mathbf{x}}
\newcommand{\yy}{\mathbf{y}}

\newcommand{\vQ}{\mathbf{Q}}
\newcommand{\vK}{\mathbf{K}}
\newcommand{\vA}{\mathbf{A}}
\newcommand{\vB}{\mathbf{B}}

\newcommand{\calX}{\mathcal{X}}

\newcommand{\calH}{\mathcal{H}}

\newcommand{\R}{\mathbb{R}}

\newcommand{\BERT}{\mathrm{BERT}}
\newcommand{\ELMO}{\mathrm{ELMo}}

\everypar{\looseness=-1}

\title{A Non-Linear Structural Probe}

\usepackage{tipa}
\newcommand{\ucambridge}{\normalfont \text{\textipa{D}}}
\newcommand{\ethz}{\text{\normalfont \textipa{Q}}}
\newcommand{\uedinburgh}{\normalfont \text{\textipa{@}}}

\author{
Jennifer C. White\raise1.0ex\hbox{\normalfont\ucambridge}~\;~ 
Tiago Pimentel\raise1.0ex\hbox{\normalfont\ucambridge}~\;~ Naomi Saphra\raise1.0ex\hbox{\normalfont\uedinburgh}~\;~ Ryan Cotterell\raise1.0ex\hbox{\normalfont\ucambridge\!,\ethz}\\
  \raise1.0ex\hbox{\normalfont\ucambridge}University of Cambridge,~\;~ \raise1.0ex\hbox{\normalfont\uedinburgh}University of Edinburgh,~\;~ \raise1.0ex\hbox{\normalfont\ethz}ETH Z\"{u}rich \\ \\
  \texttt{jw2088@cam.ac.uk},~\;~ \texttt{tp472@cam.ac.uk} \\ \texttt{n.saphra@ed.ac.uk},~\;~ \texttt{ryan.cotterell@inf.ethz.ch}
}

\date{}

\begin{document}
\maketitle
\begin{abstract}
Probes are models devised to investigate the encoding of knowledge---e.g. syntactic structure---in contextual representations.
Probes are often designed for simplicity, which has led to restrictions on probe design that may not allow for the full exploitation of the structure of encoded information; one such restriction is linearity.
We examine the case of a structural probe \cite{hewitt-manning-2019-structural}, which aims to investigate the encoding of syntactic structure in contextual representations through learning only linear transformations.
By observing that the structural probe learns a metric, we are able to kernelize it and develop a novel non-linear variant with an identical number of parameters.
We test on 6 languages and find that the radial-basis function (RBF) kernel, in conjunction with regularization, achieves a statistically significant improvement over the baseline in all languages---implying that at least part of the syntactic knowledge is encoded non-linearly.
We conclude by discussing how the RBF kernel resembles $\BERT$'s self-attention layers and speculate that this resemblance leads to the RBF-based probe's stronger performance.
\end{abstract}

\section{Introduction}
Probing has been widely used in an effort to better understand what linguistic knowledge may be encoded in contextual word representations such as $\BERT$ \cite{devlin2018bert} and $\ELMO$ \cite{peters-etal-2018-deep}.
These probes tend to be designed with simplicity in mind and with the intent of revealing what linguistic structure is encoded in an embedding, rather than simply learning to perform an NLP task \citep{hewitt-liang-2019-designing,zhang-bowman-2018-language,voita-titov-2020-information}
This preference for simplicity has often led researchers to place restrictions on probe designs that may not allow them to fully exploit the structure in which information is encoded \citep{saphra-lopez-2019-understanding,pimentel-etal-2020-information,pimentel-etal-2020-pareto}.
This preference has led many researchers to advocate the use of linear probes over non-linear ones \cite{alain2016understanding}.

This paper treats and expands upon the structural probe of \newcite{hewitt-manning-2019-structural}, who crafted a custom probe with the aim of investigating the encoding of sentence syntax in contextual representations.
They treat probing for syntax as a distance learning problem: they learn a linear transformation that warps the space such that two words that are syntactically close to one another (in terms of distance in a dependency tree) should have contextual representations whose Euclidean distance is small.
This linear approach performs well, but the restriction to learning only linear transformations seems arbitrary.
Why should it be the case that this information would be encoded linearly within the representations?\looseness=-1

In this paper, we recast \newcite{hewitt-manning-2019-structural}'s structural probing framework as a general metric learning problem. This reduction allows us to take advantage of a wide variety of non-linear extensions---based on kernelization---proposed in the metric learning literature \cite{kulis2013metric}.
These extensions lead to probes with the \emph{same} number of parameters, but with an increased expressivity.

By exploiting a kernelized extension, we are able to directly test whether a structural probe that is capable of learning non-linear transformations improves performance. 
Empirically, we do find that non-linearity helps---a structural probe based on a radial-basis function (RBF) kernel improves performance significantly in all 6 languages tested over a linear structural probe. 
We then perform an analysis of $\BERT$'s attention, asserting it is a rough approximation to an RBF kernel.
As such, it is not surprising that the syntactic information in $\BERT$ representations is more accessible with this specific \emph{non-linear} transformation. 
We conclude that kernelization is a useful tool for analyzing contextual representations---enabling us to run controlled experiments and investigate the structure in which information is encoded.

\section{The Structural Probe}
\newcite{hewitt-manning-2019-structural} introduce the \defn{structural probe}, a novel model designed to probe for syntax in contextual word representations. We review
their formulation here and build upon it in \cref{sec:metric-learning}.
A sentence $\vw$ lives in a space $V^*$, defined here as the Kleene closure of a (potentially open) vocabulary $V$. 
The syntactic distance $\Delta_{ij}$ between any two words in a sentence $\vw$ is the number of steps needed to go from one word to the other while walking in the sentence's syntactic tree. 
More formally, if we have
a dependency tree $\vt$ (a tree on $n+1$ nodes) of a sentence $\vw$ of length $n$, we define $\Delta_{ij}$ as the length
of the shortest path in $\vt$ between $w_i$ and $w_j$; this may be computed, for example, by \citeauthor{floyd1962algorithm}--\citeauthor{warshall1962theorem}. %
Contextual representations of a sentence $\vw$ are
a sequence of vectors $\vh_i \in \R^{d_1}$ that encode some linguistic knowledge about a sequence. 
In the case of $\BERT$, we have
\begin{equation}
    \vh_i = \BERT(\vw)_i \in \R^{d_1}
\end{equation}

Here, the goal of probing is to evaluate whether the contextual representations capture the syntax in a sentence. 
In the case of the structural probe, the goal is to see whether the syntactic distance between any two words can be approximated by a learned, linear distance function:
\begin{align}\label{eq:distance}
    \dB(\vh_i, \vh_j) = ||\vB\vh_i -\vB\vh_j||_2
\end{align}
where $\vB \in \R^{d_2 \times d_1}$ is a linear projection matrix.
That is to say, they seek a linear transformation such that the transformed contextual representations relate to one another roughly as their corresponding words do in the dependency tree.
To learn this probe, \citeauthor{hewitt-manning-2019-structural} minimize the following per-sentence objective with respect to $\vB$ through stochastic gradient descent
\begin{equation}\label{eq:min_distance}
    \frac{1}{|\vw|^2}\sum_{i=1}^{|\vw|} \sum_{j=i+1}^{|\vw|} |\Delta_{ij} - \dB(\vh_i, \vh_j)|
\end{equation}
This is simply minimizing the difference between the syntactic distances obtained from the dependency tree and the distance between the two vectors under our learned transformation.
From the pairwise distances predicted by the probe, \citeposs{prim1957shortest} algorithm can be used to recover the one-best undirected dependency tree.

\section{Kernelized Metric Learning}
The restriction to a linear transformation may hinder us from uncovering some of the syntactic structure encoded in the contextual representations. Indeed, there is no reason \textit{a-priori} to expect that $\BERT$ encodes its knowledge in a fashion that is specifically accessible to a \emph{linear} model.
However, if we were to introduce non-linearity by using a neural probe, for example, we would have to pit a model with very few parameters (the linear model) against one with very many (the neural network); this comparison is not fair and also goes against
the spirit of designing simple probes. To preclude %
the need for a neural probe, we instead turn to a kernelized probe.

The key insight is that the structural probe reduces the problem of probing for linguistic structure to that of \defn{metric learning} \cite{kulis2013metric}.
This can be clearly seen in \cref{eq:min_distance}, where the probe learns a distance metric between two representations in such a way that it matches the syntactic one.
Recognizing this relationship allows us to take advantage of established techniques from the metric learning literature to improve the performance of the probe without increasing its complexity, e.g. through kernelization. 

\subsection{The ``Kernel Trick'' for Distances}

Many algorithms in machine learning, e.g. support vector machines and $k$-means, can be \defn{kernelized} \cite{scholkopf2002learning},
thus allowing for linear models to be adapted into non-linear ones.
Expanding on a classic result \cite{schoenberg1938metric}%
, \newcite{scholkopf2001kernel} show that any positive semi-definite (PSD) kernel can be used to construct a distance in a Hilbert space $\calH$. 
Formally, their result states that for any PSD kernel $\kappa: \calX \times \calX \rightarrow \R_{\geq 0}$, there exists a feature map $\phi: \calX \rightarrow \calH$ such that
\begin{align}
    ||\phi(\xx) - \phi(\yy)&||_2 = \\
    &\sqrt{\kappa(\xx, \xx) - 2\kappa(\xx, \yy) + \kappa(\yy, \yy)} \nonumber
\end{align}
This generalizes \cref{eq:distance} to yield
a new, non-linear distance metric.
This means that we can achieve the effects of using some non-linear feature mapping $\phi$ without %
having to specify it: we need only specify a kernel function and perform calculations using this kernelized distance metric. 
Importantly, as opposed to deep neural probes, this learnable metric
has an \emph{identical} number of parameters to the original.%
\footnote{We note that we do not use selectivity \citep{hewitt-liang-2019-designing} to control for probe complexity since it does not apply to this syntax tree reconstruction task---selectivity control tasks work at the word type level, as opposed to the sentence one.}

\subsection{Common Kernels}
In this section we introduce the kernels to be used.
These kernels were chosen as they represent a comprehensive selection of commonly-used kernels in the metric learning literature \cite{kulis2013metric}.
The original work of \newcite{hewitt-manning-2019-structural} makes use of the \defn{linear kernel}:
\begin{equation}
    \kappa_{\mathrm{linear}}(\vh_i, \vh_j) = (\vB\vh_i)^{\top}(\vB\vh_j)
\end{equation}
The first non-linear kernel we consider is the  \defn{polynomial kernel}, defined as
\begin{equation}
    \kappa_{\mathrm{poly}}(\vh_i, \vh_j) = \left((\vB\vh_i)^{\top}(\vB\vh_j) + c\right)^d
\end{equation}
where $d \in \mathbb{Z}_{+}$ and $c \in \R_{\geq 0}$. A polynomial kernel of degree $d$ allows for $d$-order interactions between the terms. 
When working with $\BERT$, this means that we may construct $d$-order conjunctions of the dimensions of the contextual representations input into the probe.
Next, we consider the \defn{radial-basis function kernel} (RBF).
This kernel is also called the Gaussian kernel
and is defined as
\begin{equation}
    \kappa_{\mathrm{rbf}}(\vh_i, \vh_j) = \exp \left(- \frac{||\vB\vh_i -\vB\vh_j||^2}{2 \sigma^2} \right)
\end{equation}
This kernel has an alternative interpretation as a similarity measure between both vectors, being at its maximum value of $1$ when $\vh_i = \vh_j$.
In contrast to the polynomial kernel, the Gaussian kernel implies a feature map in an infinite dimensional Hilbert space.
When the RBF kernel is used in our probe, we may rewrite \cref{eq:distance}
as follows:
\begin{align} \label{eq:distance_rbf}
      &d_{\kappa_{\mathrm{rbf}}}(\vh_i, \vh_j)^2 \\
      &= \kappa_{\mathrm{rbf}}(\vh_i, \vh_i) - 2\kappa_{\mathrm{rbf}}(\vh_i, \vh_j) + \kappa_{\mathrm{rbf}}(\vh_j, \vh_j)  \nonumber \\
      &= 2 - 2\kappa_{\mathrm{rbf}}(\vh_i, \vh_j) \nonumber \\
      &= 2 - 2 \exp \left(- \frac{||\vB\vh_i -\vB\vh_j||^2}{2 \sigma^2} \right) \nonumber
\end{align}
Which is similar to the original linear case in \cref{eq:distance}, but with a scaling term $-\frac{1}{2\sigma^2}$ and a non-linearity $\exp(\cdot)$. Finally, we consider, the \defn{sigmoid kernel}, which is defined as
\footnote{\newcite{lin2003study} observe that it is difficult to effectively tune $a$ and $b$ in the sigmoid kernel. They also note that although this kernel is not in fact PSD, it is PSD when $a$ and $b$ are both positive, which we enforce in this work.}
\begin{equation}
    \kappa_{\mathrm{sig}}(\vh_i, \vh_j) = \tanh{(a (\vB\vh_i)^{\top}(\vB\vh_j) + b)} 
\end{equation}
\section{Regularized Metric Learning}\label{sec:metric-learning}

We also take advantage of two common regularization techniques employed in the metric learning literature to further improve the transformations learned;
both act on the matrix $\vA=\vB^{\top} \vB$ and are added to the objective
specified in \cref{eq:min_distance}.
The \defn{Frobenius norm regularizer} takes the form
\begin{equation}
    r(\vA) = ||\vA||^2_{\mathrm{F}} = \mathrm{tr}\left(\vA^{\top} \vA\right)
\end{equation}
This is the matrix analogue of the $L_2$ squared regularizer.
Minimizing the Frobenius norm of the learned matrix has the effect of keeping the values in the matrix small.
It has been a popular choice for regularization in metric learning with adaptations to a variety of problems \citep{schultz2004learning, kwok2003learning}.
We also consider the \defn{trace norm regularizer}, which is of the form
\begin{equation}
    r(\vA) = \mathrm{tr}(\vA)
\end{equation}
The trace norm regularizer is the matrix analogue of the $L_1$ regularizer and it encourages the matrix $\vA$ to be low rank.
As \newcite{jain2010inductive} point out, using a low-rank transformation in conjunction with a kernel corresponds to a supervised kernel dimensionality reduction method.
\begin{table*}
\centering
\resizebox{\textwidth}{!}{%
\begin{tabular}{lcccccccccccc}\toprule
 \multicolumn{1}{c}{} & \multicolumn{2}{c}{Basque} &
\multicolumn{2}{c}{English} &
\multicolumn{2}{c}{Finnish} &
\multicolumn{2}{c}{Korean} &
\multicolumn{2}{c}{Tamil} &
\multicolumn{2}{c}{Turkish} \\ \cmidrule(lr{.5em}){2-3} \cmidrule(lr{.5em}){4-5} \cmidrule(lr{.5em}){6-7} \cmidrule(lr{.5em}){8-9} \cmidrule(lr{.5em}){10-11} \cmidrule(lr{.5em}){12-13}
Kernel & UUAS & DSpr & UUAS & DSpr & UUAS & DSpr & UUAS & DSpr & UUAS & DSpr & UUAS & DSpr
\\ \midrule
None & 58.39 &  0.6737  
& 57.96  & 0.7382  
& 59.90 & \textbf{0.7560}  
& 68.63 &  \textbf{0.7026} 
& 48.52 & 0.5116  
& 58.87 & 0.6784  \\
Polynomial & 50.10  & 0.5751
& 59.67 &  \textbf{0.7635} 
& 57.12 &  0.7401 
& 67.58 &  0.6966 
& 54.43 &  \textbf{0.5776} 
& 55.29 &  0.6421 \\
Sigmoid & 43.14 & 0.4500  
& 41.62 & 0.6152  
& 53.14 & 0.6201 
& 44.48 & 0.3734 
& 44.30 & 0.3836 
& 51.77 & 0.5557 \\
  RBF & \textbf{60.99} & \textbf{0.6937} 
  & \textbf{62.77} & 0.7213 
& \textbf{63.08} & 0.7382 
& \textbf{71.87} & 0.6918 
& \textbf{56.96} & 0.5379  
& \textbf{61.67} & \textbf{0.6841} \\ 
\bottomrule
\end{tabular}
}
\caption{
Results of probes using various kernels, in terms of UUAS and DSpr
}
\label{table:ker_results}
\end{table*}

\section{Experiments}

We experiment with \citeposs{hewitt-manning-2019-structural} probe on 6 typologically diverse languages, following the experimental design of \newcite{maudslay-etal-2020-tale}.
Our data comes from the Universal Dependency 2.4 Treebank \cite{ud-2.4}, providing sentences and their dependency trees, annotated using the Universal Dependencies annotation scheme.%
\footnote{It was recently demonstrated by \newcite{kuznetsov-gurevych-2020-matter} that choice of linguistic formalism may have an impact on probing results. In this work, we investigate using only one formalism, so we cannot be sure that our results would not differ if an alternative formalism were used. Nonetheless, we believe that the results that we find most interesting, which are discussed in \cref{sec:rbf}, should be robust to a change in formalism, since their explanation lies in the way attention is calculated in the transformer architecture.}
For each sentence we calculate contextual representations using multilingual $\BERT$.
For all languages, we took the first 12,000 sentences (or the maximum number thereof) in the train portion of the treebank and created new 80--10--10 train--test--dev splits.\footnote{We cap the maximum number of sentences analyzed as a na\"{i}ve control for our multilingual analysis.}

We present the results from our comparison of a re-implementation of \citeposs{hewitt-manning-2019-structural} linear structural probe and the non-linear kernelized probes in \cref{table:ker_results}.
The two evaluation metrics shown are unlabeled undirected attachment score (UUAS) and the Spearman rank-order correlation (DSpr) between predicted distances and gold standard pairwise distances.
UUAS is a standard parsing metric expressing the percentage of correct attachments in the dependency tree, while
DSpr is a measure of how accurately the probe predicts the overall ordering of distances between words.
We can see that the use of an RBF kernel 
results in a statistically significant improvement in performance, as measured by UUAS, in all 6 of the languages tested.\footnote{Significance was established using paired permutation tests with 10,000 samples, to the level of $p<0.05$.} 
For some languages this improvement is quite substantial, with Tamil seeing an improvement of 8.44 UUAS from the baseline probe to the RBF kernel probe.
\section{The RBF Kernel and Self-Attention}\label{sec:rbf}
The RBF kernel produces improvements across all analyzed languages.
This suggests that it is indeed the case that syntactic structure is encoded non-linearly in $\BERT$.
As such, analyzing this specific kernel may yield insights into what this structure is.
Indeed, none of the other kernels systematically improve over
the linear baseline, implying this is not just an effect of the non-linearity introduced through use of a kernel---the specific structure of the RBF kernel must be responsible.
In this section, we argue that the reason that the RBF kernel serves as such a boon to probing is that it resembles $\BERT$'s attention mechanism; recall that $\BERT$'s attention mechanism is defined as%
\looseness=-1
\begin{equation}\label{eq:attention}
    \mathrm{att}(\vh_i, \vh_j) \propto \exp\left(\frac{(\vK \vh_i)^\top (\vQ \vh_j)}{\sqrt{d_2}}\right)
\end{equation}
where $\vK$ and $\vQ$ are linear transformations and  $d_2$ is the dimension vectors are projected into. $\vK$ projects vector $\vh_i$ into a key vector, while $\vQ$ projects $\vh_j$ into a query one. 
When the key and query vectors are similar (i.e. have a high dot product), the value of this equation is large and word $j$ attends to word $i$.

This bears a striking resemblance to the Gaussian kernel. 
Indeed, if we assume the linearly transformed representations have unit norm, i.e.  
\begin{equation}
    ||\vB\vh_i||^2 = ||\vB\vh_j||^2 = 1
\end{equation}
then we have
\begin{align}\label{eq:rbf_attention}
    \exp\bigg(&\frac{-1}{2 \sqrt{d_2}}||\vB\vh_i - \vB\vh_j||^2\bigg) \\
    &= \exp\left(\frac{-1}{\sqrt{d_2}} + \frac{(\vB\vh_i)^{\top}(\vB\vh_j)}{\sqrt{d_2}}\right) \nonumber \\
    &\propto \exp\left(\frac{(\vB\vh_i)^{\top}(\vB\vh_j)}{\sqrt{d_2}}\right) \nonumber
\end{align}
where we take $\sigma^2 = \sqrt{d_2}$. 
The similarity between \cref{eq:rbf_attention,eq:attention} suggests the attention mechanism in $\BERT$ is, up to a multiplicative factor, roughly equivalent to an RBF kernel---as such, it is not surprising that the RBF kernel produces the strongest results.\looseness=-1

The resemblance between these equations, taken together with the significant improvements in capturing syntactic distance, suggest that this encoded information indeed lives in an RBF-like space in $\BERT$.
Such information can then be used in its self-attention mechanism; allowing $\BERT$ to pay attention to syntactically close words when solving the cloze language modeling task.
Being attentive to syntactically close words would also be supported by recent linguistic research, since words sharing syntactic dependencies have higher mutual information on average \cite{futrell-etal-2019-syntactic}.

The representations we analyze, though, are taken from $\BERT$'s final layer; as such, they are not trained to be used in any self-attention layer---so why should such a resemblance be relevant?
$\BERT$'s architecture is based on the Transformer \cite{vaswani2017attention}, and uses skip connections between each self-attention layer.
Such skip connections create an incentive for residual learning, i.e. only learning residual differences in each layer, while propagating the bulk of the information \cite{he2016deep}. 
As such, $\BERT$'s final hidden representations should roughly live in the same manifold as its internal ones.

It is interesting to note that the RBF kernel achieves the best performance in terms of UUAS in all languages, but it only twice achieves the best performance in terms of DSpr.
This may be due to the fact that, as we can see by examination of \cref{eq:distance_rbf}, the distance returned by the RBF kernel will not exceed 2, whereas syntactic distances in the tree will.
Further, the gradient of the RBF kernel contains an exponential term which will cause it to go to zero as distance increases (while an examination of the unkernelized loss function reveals the opposite behavior).
This means that it will be less sensitive to the distances between syntactically distant words and focus more on words with small distances.
This may partially explain its better performance on UUAS, and comparably worse performance as measured by correlation (which counts pairwise differences between \emph{all} words, not just those which are directly attached in the tree). 
Furthermore, our probe's focus on nearby words resembles the general attentional bias towards syntactically close words \citep{voita_analyzing_2019}.

The direct resemblance between self-attention mechanisms and our proposed probe metric poses a new way of understanding results from more complex probes. 
While \citet{reif_visualizing_2019} understood the Euclidean-squared distance of \citeauthor{hewitt-manning-2019-structural} as an isometric tree embedding, their geometric interpretation did not factor in the rest of BERT's architecture. Such simplified contextless probes cannot tell us how linguistic properties are processed by a sequence of learned modules~\citep{saphra-lopez-2019-understanding}. 
However, we consider representations \emph{in the context} of the model which is expected to employ them. From this perspective, simpler metrics may be rough approximations to our RBF kernel space, which is actually capable of measuring linguistic properties that can be easily extracted by an attention-based architecture.\looseness=-1

\section{Conclusion}
We find that the linear structural probe \cite{hewitt-manning-2019-structural} used to investigate the encoding of syntactic structure in contextual representations can be improved through kernelization, yielding a non-linear model.
This kernelization does not introduce additional parameters and thus does not increase the complexity of the probe---at least if one treats the number of parameters as a good proxy for model complexity.
At the same time, the RBF kernel improves probe performance in all languages under consideration.
This suggests that syntactic information may be encoded non-linearly in the representations produced by $\BERT$.
We hypothesize that this is true due to the similarity of the RBF kernel and $\BERT$'s self-attention layers.

\section*{Ethical Considerations}
The authors foresee no ethical concerns with the research presented in this paper. 
\bibliography{emnlp2020}
\bibliographystyle{acl_natbib}

\end{document}